# EPICURE :Ensemble Pretrained Models for Extracting Cancer Mutations from Literature

1st Jiarun Cao *†,
*National Centre for Text Mining
†dept. of Computer Science
The University of Manchester
Manchester, United Kingdom
jiarun.cao@manchester.ac.uk

2nd Elke M van Veen*†,
*Manchester Centre for Genomic Medicine
St Mary's Hospital
Manchester University Hospitals NHS Foundation Trust
†Division of Evolution and Genomic Sciences
School of Biological Sciences
Faculty of Biology, Medicine and Health
The University of Manchester
Manchester, United Kingdom
elke.vanveen@manchester.ac.uk

3rd Niels Peek *†,
*Division of Informatics,
Imaging and Data Science
†NIHR Manchester Biomedical
Research Centre
Faculty of Biology, Medicine and Health
The University of Manchester
Manchester, United Kingdom
niels.peek@manchester.ac.uk

4th Andrew G Renehan *†,
*Division of Cancer Sciences
School of Medical Sciences
Faculty of Biology, Medicine and Health
The University of Manchester
†Surgery, Christie NHS Foundation Trust
Manchester, United Kingdom
andrew.renehan@manchester.ac.uk

5th Sophia Ananiadou*†,
*National Centre for Text Mining
†dept. of Computer Science
The University of Manchester
Manchester, United Kingdom
sophia.ananiadou@manchester.ac.uk

*Abstract*—To interpret the genetic profile present in a patient sample, it is necessary to know which mutations have important roles in the development of the corresponding cancer type. Named entity recognition (NER) is a core step in the text mining pipeline which facilitates mining valuable cancer information from the scientific literature. However, due to the scarcity of related datasets, previous NER attempts in this domain either suffer from low performance when deep learning based models are deployed, or they apply feature-based machine learning models or rule-based models to tackle this problem, which requires intensive efforts from domain experts, and limit the model generalization capability. In this paper, we propose EPICURE, an ensemble pre-trained model equipped with a conditional random field pattern (CRF) layer and a span prediction pattern (Span) layer to extract cancer mutations from text. We also adopt a data augmentation strategy to expand our training set from multiple datasets. Experimental results on three benchmark datasets show competitive results compared to the baseline models, validating our model's effectiveness and advances in generalization capability.

*Index Terms*—biomedical text mining, named entity recognition, pre-trained model, cancer mutation detection

## I. INTRODUCTION

Over the past ten years, a large number of variants have been associated with an increased risk of cancer disease, rather than being directly causative for the development of cancer. For these variants associated with an increased risk, there is currently a lack of understanding of the underlying biological mechanism leading to cancer disease susceptibility [1]. Therefore, genetic association studies in cancer are used to identify candidate genes or genome regions that contribute to the risk of acquiring a specific cancer by testing for a correlation between disease onset and genetic variation. Such studies have revealed that many diseases are the result of susceptibility mutation in many different genes. The combined risk associated with a large number of different genetic variations is then typically expressed as a polygenic risk score [2], [3]. The main shortcoming of genetic association studies is that they cannot tell us through which mechanism susceptibility mutations lead to disease onset: they can only tell us that "something is going on". Once a mutation has been identified as associated with cancer disease risk, researchers have to search the scientific literature of plausible biologic mechanisms that explain the association – and this is typically a slow and labour-intensive process [4].

In this paper, we suggest methods for extracting mutation-disease associations found in genetic studies by automatically mining the scientific literature [5]. To be more specific, we use named entity recognition(NER) methods to extract information relevant for cancer mutation detection tasks [6]–[8]. However, due to the limited availability of existing cancer mutation datasets, commonly-used deep learning methods [10]–[13], [19], i.e. CNN-CRF method [7], BiLSTM-CNN based model [10], BiLSTM-based CRF models [11]–[13] do not perform well in cancer mutation tasks since these models rely on massive volume of training data.

To address the limited scale of the dataset [6], [10], [20], [21], part of the mainstream existing methods focused on using rule-based methods to match the eligible mutation entity in the text [1], [6], [7], [10], but this approach cannot be generalized in other datasets since it is specifically crafted for a certain dataset. Others use machine learning approaches, which requires feature engineering for training models [16], [18], [21], [23].

We propose an ensemble pre-trained model that combines a CRF pattern and a span-prediction pattern to automatically extract the mutations from cancer literature without depending on any hand-crafted linguistic features or rules. We first design two model patterns, where the first one predicts mutation entities in a token-by-token manner with a CRF layer [22], and the second identifies mutation entities by predicting the entire possible spans with a pointerNet [32]. Then we introduce the pre-trained model as our model encoder to leverage prior knowledge from massive pre-trained corpora. Furthermore, we also propose a strategy to expand our training set across several homogeneous datasets. Experimental results show that our model can achieve comparable performance to existing methods in three well-studied datasets without involving any external resources.

The contribution of our paper can be summarised as follows:

- We propose an ensemble pre-trained extractor, which consists of a combined CRF and a span prediction pattern to automate the process of identifying mutation entities from cancer literature without using external resources. We also adopt a data augmentation strategy to expand our training set by collecting valid training samples from multiple datasets.
- Experimental results show that the proposed model can achieve comparable results in three datasets compared to existing baseline models, which verifies our model effectiveness, as well as the advance in model generalization capability.

## II. Related Work

To obtain key information for personalized medicine and cancer research, clinicians and researchers in the biomedical domain are in great need of searching genomic variant information from the biomedical literature now more than ever before. Due to the various written forms of genomic variants, however, it is difficult to locate the right information from the literature when using a general literature search system. To address the difficulty of locating genomic variant information from the literature, researchers have suggested various solutions based on automated literature-mining techniques [15]–[18], [21], [22]. There are several genomic variant NER tools that are publicly available [1], [6], [7], [10], [17], [25], [26], [28]–[30]. Most of the tools are based on a set of regular expressions to recognize the genomic variants, while others are machine learning-based. We briefly describe the most prevalent tools for genomic mutation extraction.

MutationFinder [1] is a regular expression-based mutation NER tool. Based on the six rules that the authors found to yield good precision and recall of recognizing mutations from the text. EMU [6] is another regular expression-based mutation extraction tool that can find not only point mutations but also insertion or deletion mutations. After finding mutations, EMU uses another set of regular expressions to filter out false-positive mutation terms. SETH [7] supports both named entity recognition and named entity normalization for genomic variants. First, for NER of genomic variants, SETH uses extended BackusNaur form and the regular expressions from MutationFinder together to find genomic variants. tmVar [10] uses CRF with regular expression to find genomic variants and to map each variant with a related gene that is recognized using the GnormPlus NER tool [31]. tmVar also supports named entity normalization, which normalizes the recognized gene and genomic variant pair into RSID.

AVADA [17] employs machine learning and regular expressions to improve gene-variant mapping in full-text articles. It uses 47 regular expressions to recognize genomic variants. It also employs a custom-built gene name recognition tool that learns and uses lists of gene names obtained from the HUGO Gene Nomenclature Committee (HGNC) and the UniProt database for finding candidate gene names associated with variants in an article. MEMA [25] expanded the multiple regular expressions, as used by MutationFinder, to recognize 30 other patterns of mutations that are not the "wild-type amino acid–location–mutated type of amino acid" form. MEMA finds the gene name for each mutation, using the HUGO database as a dictionary, and finds gene names from the abstract or a sentence in which that mutation is mentioned. Nala [13] utilized CRF + CBOW word embedding as model structure along with BIOE tagging scheme. It extracts both target standard(ST) mutation mentions and natural language(NL) mentions at the same time. Nala captured NL and ST by combining conditional random fields with word embedding features learned unsupervised from the entire PubMed. DiMeX [26] gathers such mutation-disease associations into public knowledge bases. DiMeX modules including syntactic preprocessing of input text, detecting different types of mutations, a novel algorithm to associate mutations with genes, an information extraction (IE) module to apply lexical and semantic patterns to extract associations between mutations and diseases, and additional rules to infer associations beyond patterns. The mutation extraction employs a list of regular expression patterns to detect mutations containing the three components, wild-type symbol, mutant-type symbol, and the position. Similar to EMU, there are several other tools that use sequence filtering to find gene names in which the target mutation resides. MuteXt [9] and MutationGraB [28] are early-stage, single-point mutation-extraction tools that use regular expressions. After finding mutation names using regular expressions, these tools also find candidate gene names in the text around the mutation and use an amino acid sequence to filter the gene names, as does EMU.

## III. Methods

Our method of identifying cancer mutations is depicted in Figure 1 and 2, and it can be summarised as follows. Firstly,

we propose a method for expanding our training set from multiple corpora. Then, we illustrate the model structure, of which it constitutes a BERT-based [14] encoder layer and two separate model patterns in the upper layer denoting two different methods of identifying mutations. Finally, the model resulting from two CRF and span patterns can be incorporated using majority voting. In the following sections, we provide a more detailed description of this multi-step approach.

### A. Expanding training set

Existing methods [10], [12], [15] are typically trained on a single corpus that is fully annotated with a uniform set of semantic types. For multiple corpora, this results in multiple models, each of which covers only a small slice of the semantic space. In terms of cancer mutation corpora, the scale of each dataset is insufficient to train a deep learning model. However, directly copying samples from other dataset can lead to excessive negative instances, which can further deteriorate the model performance.

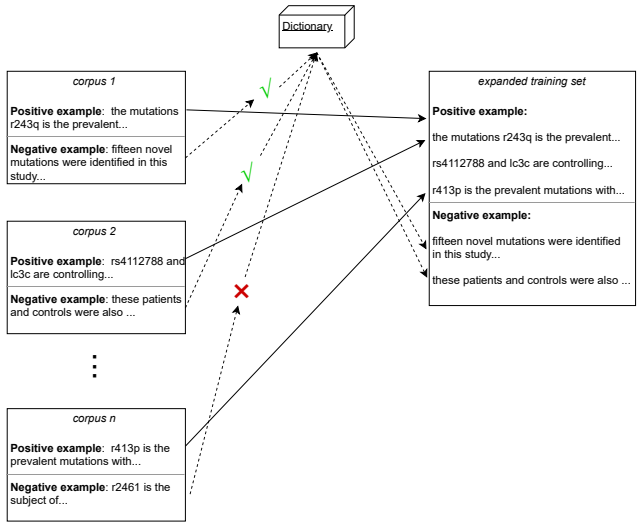

Fig. 1: Workflow of filtering and expanding the training set

To avoid such negative effects from the inconsistent annotation of types that are not shared across dataset, whilst also benefiting from the annotation of shared types between training data, it is necessary to generate from each dataset only those training instances that are relevant to types annotated in that dataset. In the setting considered here, for positive instances, we leave them in the expanded training set since positive instances are always explicitly annotated. On the other hand, for generation of negative instances, we limit the generation of candidate negative instances in each dataset of the merged training set to only those cases in which the surface expression (or, e.g., its base form) matches at least one positive instance of an annotated type in any dataset that shares the type [27]. As shown in Figure 1, we first construct a dictionary storing all the mutation mentions across the datasets, then we only select examples of which at least one token appears in the dictionary as our negative samples. In this way, we can construct an expanded training set containing valid positive and negative examples over all datasets.

### B. Model Encoder

Although deep learning-based methods [10]–[13], [19] have been successfully applied to biomedical named entity recognition, they required pre-trained word embeddings that were often learned from a large corpus of unannotated texts. Specifically in cancer mutation detection tasks, the deep learning models struggle to capture enough valid information from literature due to the limited scale of datasets. In this study, we propose a pipeline equipped with the pre-trained model. We test three different BERT-based pre-trained models [14], [23], [33]for the fine-tuning procedure, respectively.

To be more specific, BERT was trained on Wikipedia and BooksCorpus. BioBERT was initialized with *BERT_based_cased* model and pre-trained with additional biomedical corpus including PubMed abstracts (PubMed), PubMed Central full-text articles (PMC), or PubMed+PMC. SciBERT is a pre-trained model for scientific text based on BERT. They train SCIBERT on a random sample of 1.14M papers from Semantic Scholar. As shown in figure 2, BERT-based pre-trained models leverage a special token [CLS] to aggregate the whole representation of the input text and a special token [SEP] to indicate the end of the input text. Therefore, given a input text $X = x1, x2, ..., xn$, we firstly add special tokens in the start and end of input text, then we use BERT-based pre-trained models to encode the input text $X$ into contextual word representation $H$:

$$X^{'} = [CLS], x1, x2, ..., xn, [SEP] \quad (1)$$

$$H = BERT_{finetune}(X^{'}) \quad (2)$$

Where $BERT_{finetune}$ denotes that it will be fine-tuned during training. Therefore, $BERT_{finetune}$ will learn a corresponding generalization of sentence representations and adapt to our own training set.

### C. Mutation Recognition

In this section, we propose two different patterns to tackle the cancer mutation recognition problem, and then do model ensemble using majority voting strategy.

*a) CRF pattern.:* In the first pattern, we consider mutation recognition problem as a sequence-labeling task. Similar to TmVar [10], each mutation component was considered as an individual label such that every mutation mentioned becomes a sequence of labels. We assign two different labeling methods: BIO(Beginning, Inside, Outside) and BMEO(Beginning, Middle, End, Outside). Accordingly, we then adapt a probability-based sequence detection CRF model [22], which defines the conditional probability distribution $P(Y|H)$ of label sequence $Y$ given contextual word representation $H$ aforementioned.

$$P(Y|H) = \frac{exp(H, Y)}{\sum_{Y'} exp(F, (H, Y'))} \quad (3)$$

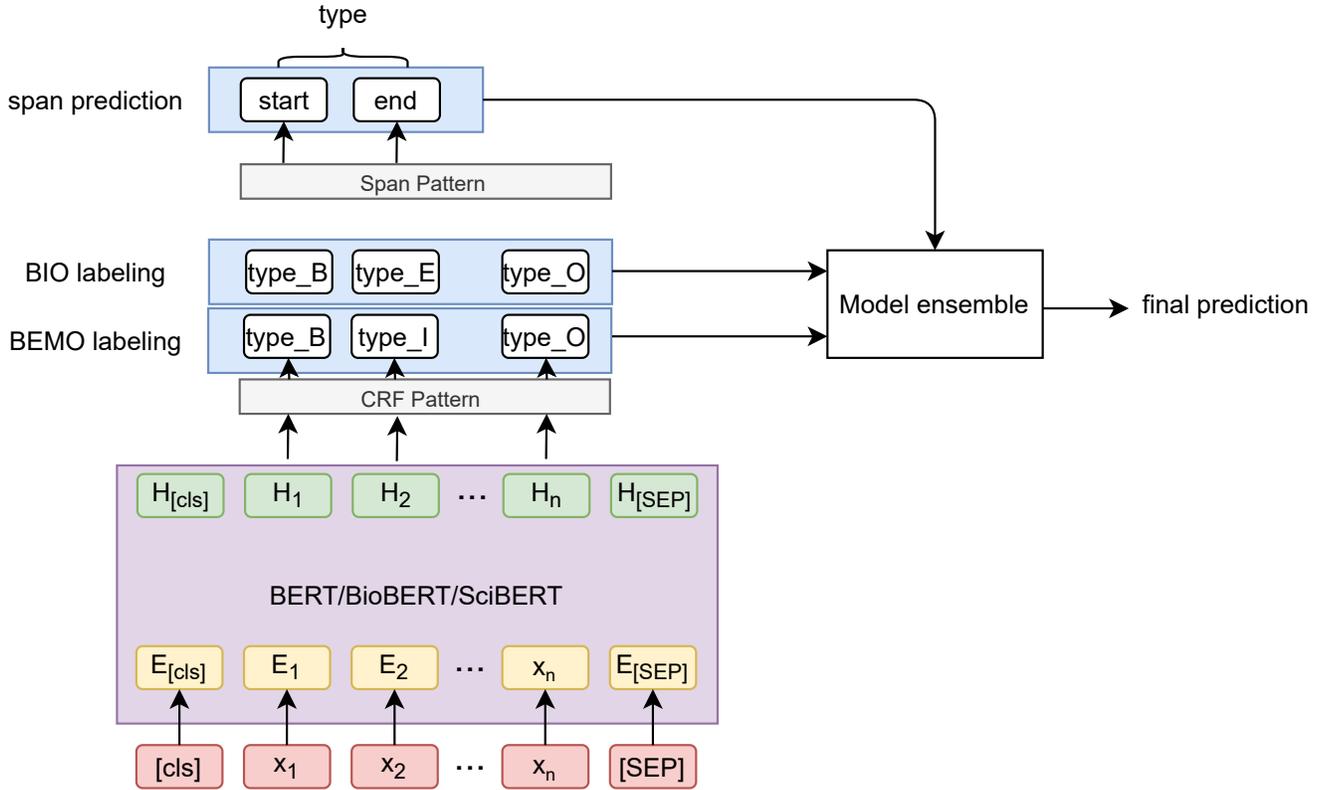

Fig. 2: Overview of model architecture

where $y_1, ..., y_n$ is a label sequence from $Y$. $F(H, Y) = \sum_{j=1}^{n} \sum_{i=1}^{w} \omega_i f_i(y_j, y_{j-1}, X)$ is a global feature vector for label sequence $Y$ and contextual word representation $H$ and and $\omega_1, ..., \omega_w$ is a feature weight vector.

*b) Span pattern.:* In the second pattern, instead of assigning labels to each token, we use two fully-connected layer named starting layer and ending layer to predict the mutation labels upon each tokens given contextual word representation $H$:

$$P_{starting}(Y_i|H_i) = \frac{exp(H_i)}{\sum_J exp(H_j)} \quad (4)$$

$$P_{ending}(Y_i|H_i) = \frac{exp(H_i)}{\sum_J exp(H_j)} \quad (5)$$

Where $H_i$ denotes the current contextual word representation in sequence $H$, $j \in J$ denotes the mutation label types.

In the inference stage, at position $t$, if the mutation type $Y$ in current word is not none in the starting layer, we then look for the position $t'$ in the ending layer after position $t$, where its mutation type is the same as $Y$ in the starting layer. and the span of mutation mention is denoted as position $t'-t$ with corresponding label type $Y$.

*c) Model ensemble.:* To take advantage of both model patterns, we introduce model ensemble using majority voting. To be more specific, in terms of each token in the input text, we can obtain three predicted labels deriving from CRF pattern (BIO and BEMO) and span pattern, we convert all three labels into BIO tagging form, and select the majority type as our final prediction.

IV. EXPERIMENTS

*A. Experiment Settings*

The model encoder layer follows that of BERT-Large using a 24-layer Transformer with 1024 hidden size and 16 self-attention heads. In the fine-tuning stage, most model hyperparameters were the same as those saved in BERT [14]. We set up batch size as 24, and the number of training epochs as 100. we optimize network parameters by Adam with a 3e-5 learning rate. The dropout rate is 0.1 and weight decay is 0.01. We also set up the maximum length of the input sentence is 512.

*B. Datasets*

We evaluate our model on three well-studied datasets: TmVar [10], EMU [6] and BRONCO [24]. TmVar uses PubMed to obtain a corpus of MEDLINE abstracts that contain a large number of mutation mentions. EMU uses a set of 218 UMLS concepts denoting forms of PCa as a filter for UMLS concepts identified by MetaMap in the title and abstract of MEDLINE citations. BRONCO contains more than 400 variants and their relations with genes, diseases, drugs and cell lines in the context of cancer and anti-tumor drug screening research. The variants and their relations were manually extracted from 108 full-text articles. The detailed statistics are listed in Table I.

### C. Baselines

We select the three state-of-the-art models on each dataset as our baselines respectively.

MutationFinder [1] is a regular expression-based mutation NER tool. Based on the six rules that the authors found to yield good precision and recall of recognizing mutations from the text, MutationFinder uses over 700 regular expressions to recognize not only point mutations with wild-type residue forms but also the mutations in simple natural language forms.

EMU [6] is another regular expression-based mutation extraction tool that can find not only point mutations but also insertion or deletion mutations. After finding mutations, EMU uses another set of regular expressions to filter out false-positive mutation terms. EMU also finds the gene name with which the mutation is associated.

TmVar [10] uses conditional random fields (CRF) with regular expression to find genomic variants and to map the variant with a related gene that is recognized using the GnormPlus NER tool. tmVar also supports NEN, which normalizes the recognized gene and genomic variant pair into RSID.

## V. RESULT

### A. Comparisons of different baseline models

Table II shows the performance of existing state-of-the-art methods and our best fine-tuned models on three aforementioned datasets, the Table II shows that our models gain a comparable result considering the baseline models. Specifically, our model outperformed previous methods and achieved the state-of-the-art performance by 1.3% F1 score on BRONCO. We can notice that we achieve a relatively high recall, a possible reason is due to the differences in scope, and broader coverage of different types of mutations in the training data that we utilise. In terms of TmVar and EMU datasets, our model is 2.5% and 2.9% lower than TmVar by F1 score. It indicates that our model is able to advance without involvement of human labours. Meanwhile we can notice that existing models underperform significantly if it transfers to other datasets, however, our model can achieve relatively stable results on all three datasets where it proves the generalization capability of our model.

Table III shows the metrics of different mutation types on TmVar datasets. As we can notice that most of mutation types including `DELETION`, `DUPLICATION`, `FRAME SHIFT`, `INSERTION`, `SNP`, `SUBSTITUTION` all achieves reasonable result, while rare type, i.e., `INDEL` can not be predicted properly due to the scarce frequency of it in our training set. Therefore, the lack of ability to predict rare mutation types is also the bottleneck of our model.

### B. Comparisons of different pre-trained models

To test the compatibility of different pre-trained models in our task, we compare the different pre-trained model performance on TmVar dataset. As shown in Table IV, we see that the BioBERT-based-cased model achieves the best F1 score compared to the other pre-trained models. Therefore, we adopt the BioBERT-based-cased model in the rest of experiments.

|  | Precision | Recall | F1 score |
|---|---|---|---|
| traditional word emb. | 89.7% | 69.2% | 78.1% |
| BERT-based-uncased | 79.9% | 54.0% | 64.4% |
| BERT-based-cased | 76.3% | 81.5% | 80.7% |
| SciBERT-based-uncased | 74.8% | 83.0% | 78.7% |
| BioBERT-base-cased | 76.3% | 82.4% | 81.0% |

TABLE IV: Experiment results with different pre-trained models on TmVar dataset

### C. Ablation study

To verify the effectiveness of expanding our training set and model ensemble, we test our model component separately, Table V shows the importance of the different crucial components in our model, we can notice the model outperformance 2.2%, 0.9%, and 9.4% F1 score by expanding our training set in TmVar, BRONCO, and EMU datasets respectively. Accordingly, our model also advances 2.0%, 1.0%, 1.2% F1 score in the aforementioned datasets by doing model ensemble.

|  | TmVar | EMU | BRONCO |
|---|---|---|---|
| Raw | 86.7% | 88.9% | 85.7% |
| Model Ensemble | 88.7% | 89.9% | 86.9% |
| Expanding training set | 88.9% | 90.8% | 93.8OK % |

TABLE V: Ablation study by F1 score with different experimental settings

## VI. CONCLUSION

In this study, we developed a novel mutation entity recognition approach for the cancer domain, of which the input text is encoded by pre-trained models and then the mutation type in each token is predicted by CRF pattern and span pattern independently. To deal with the limited scale of training set, we also propose a strategy for expanding our training set, alongside model ensemble to boost the model robustness.

Compared to existing methods, our model can achieve comparable results when datasets vary, while existing methods can only perform well on the certain types of dataset.

However, our model suffers from label imbalance problems to some extent, which indicates that the model is incapable of predicting mutation types if they appear rarely in the datasets. This problem can be explored in our future work.

## VII. ACKNOWLEDGMENT

This work was supported by CRUK via the funding to Cancer Research UK Manchester Major Centre [C19941/A27859]. AGR is supported by the Manchester NIHR Biomedical Research Centre (IS-BRC-1215-20007).

|  | TmVar | EMU | BRONCO |
|---|---|---|---|
| Type | title and abstract | title and abstract | full-text |
| Contained Info | variants | variants, genes, diseases | vars, genes, diseases, drug, cell line |
| Number of Documents | 500 | 109 | 108 |
| Mutation Types | Sub, Del, Ins, Dup, InDel, SNP, FS | Sub, Del, Ins, SNP, FS | Sub, Del, Ins, InDel, SNP, FS |
| Unique Var | 871 | 172 | 275 |

TABLE I: Statistics of the evaluated datasets

|  | TmVar | | | EMU | | | BRONCO | | |
|---|---|---|---|---|---|---|---|---|---|
|  | Prec | Rec | F1 | Prec | Rec | F1 | Prec | Rec | F1 |
| Mutation Finder | 91.7% | 33.2% | 48.8% | 91.5% | 87.6% | 89.5% | 99.5% | 74.7% | 85.4% |
| EMU | 85.4% | 69.9% | 76.5% | 77.3% | 90.3% | 83.3% | 95.6% | 95.9% | 95.7% |
| TmVar | 91.4% | 91.4% | 91.4% | 76.7% | 93.8% | 84.4% | 98.8% | 95.2% | 97.0% |
| Our Model | 89.7% | 88.1% | 88.9% | 86.9% | 95.0% | 90.8% | 88.9% | 99.2% | 93.8% |

TABLE II: Model performance on different benchmark datasets

| Mutation Types | Substitution | Deletion | Insertion | Duplication | InDel | SNP | Frame Shift |
|---|---|---|---|---|---|---|---|
| Our Model | 87.8% | 82.8% | 83.9% | 80.0% | 16.7% | 78.0% | 77.8% |

TABLE III: F1 score for each mutation type on TmVar dataset.

REFERENCES

[1] Caporaso, J. Gregory, et al. "MutationFinder: a high-performance system for extracting point mutation mentions from text." Bioinformatics 23.14 (2007): 1862-1865.
[2] Duncan, L., et al. "Analysis of polygenic risk score usage and performance in diverse human populations." Nature communications 10.1 (2019): 1-9.
[3] Crouch, Daniel JM, and Walter F. Bodmer. "Polygenic inheritance, GWAS, polygenic risk scores, and the search for functional variants." Proceedings of the National Academy of Sciences 117.32 (2020): 18924-18933.
[4] Ananiadou, Sophia, Paul Thompson, and Raheel Nawaz. "" Mining events from the literature for bioinformatics applications" by S. Ananiadou, P. Thompson, and R. Nawaz; with Martin Vesely as coordinator." ACM SIGWEB Newsletter Autumn (2013): 1-12.
[5] Yepes, Antonio Jimeno, and Karin Verspoor. "Mutation extraction tools can be combined for robust recognition of genetic variants in the literature." F1000Research 3 (2014).
[6] Doughty, Emily, et al. "Toward an automatic method for extracting cancer-and other disease-related point mutations from the biomedical literature." Bioinformatics 27.3 (2011): 408-415.
[7] Thomas, Philippe, et al. "SETH detects and normalizes genetic variants in text." Bioinformatics 32.18 (2016): 2883-2885.
[8] Lee, Lawrence C., Florence Horn, and Fred E. Cohen. "Automatic extraction of protein point mutations using a graph bigram association." PLoS Comput Biol 3.2 (2007): e16.
[9] Névéol, Aurélie, W. John Wilbur, and Zhiyong Lu. "Improving links between literature and biological data with text mining: a case study with GEO, PDB and MEDLINE." Database 2012 (2012).
[10] Wei, Chih-Hsuan, et al. "tmVar 2.0: integrating genomic variant information from literature with dbSNP and ClinVar for precision medicine." Bioinformatics 34.1 (2018): 80-87.
[11] Horn, Florence, Anthony L. Lau, and Fred E. Cohen. "Automated extraction of mutation data from the literature: application of MuteXt to G protein-coupled receptors and nuclear hormone receptors." Bioinformatics 20.4 (2004): 557-568.
[12] Si, Yuqi, and Kirk Roberts. "A frame-based NLP system for cancer-related information extraction." AMIA Annual Symposium Proceedings. Vol. 2018. American Medical Informatics Association, 2018.
[13] Cejuela, Juan Miguel, et al. "nala: text mining natural language mutation mentions." Bioinformatics 33.12 (2017): 1852-1858.
[14] Devlin, Jacob, et al. "Bert: Pre-training of deep bidirectional transformers for language understanding." arXiv preprint arXiv:1810.04805 (2018).
[15] Doughty, Emily, et al. "Toward an automatic method for extracting cancer-and other disease-related point mutations from the biomedical literature." Bioinformatics 27.3 (2011): 408-415.
[16] Gui, Tao, et al. "CNN-Based Chinese NER with Lexicon Rethinking." IJCAI. 2019.
[17] Harkema, Henk, et al. "Developing a natural language processing application for measuring the quality of colonoscopy procedures." Journal of the American Medical Informatics Association 18.Supplement_1 (2011): i150-i156.
[18] Horn, Florence, Anthony L. Lau, and Fred E. Cohen. "Automated extraction of mutation data from the literature: application of MuteXt to G protein-coupled receptors and nuclear hormone receptors." Bioinformatics 20.4 (2004): 557-568.
[19] Ju, Meizhi, Makoto Miwa, and Sophia Ananiadou. "A neural layered model for nested named entity recognition." Proceedings of the 2018 Conference of the North American Chapter of the Association for Computational Linguistics: Human Language Technologies, Volume 1 (Long Papers). 2018.
[20] Mahmood, ASM Ashique, et al. "DiMeX: a text mining system for mutation-disease association extraction." PloS one 11.4 (2016): e0152725.
[21] Hu, Yuting, and Suzan Verberne. "Named Entity Recognition for Chinese biomedical patents." Proceedings of the 28th International Conference on Computational Linguistics. 2020.
[22] Lafferty, John, Andrew McCallum, and Fernando CN Pereira. "Conditional random fields: Probabilistic models for segmenting and labeling sequence data." (2001).
[23] Lee, Jinhyuk, et al. "BioBERT: a pre-trained biomedical language representation model for biomedical text mining." Bioinformatics 36.4 (2020): 1234-1240.
[24] Lee, Kyubum, et al. "BRONCO: Biomedical entity Relation ONcology COrpus for extracting gene-variant-disease-drug relations." Database 2016 (2016).
[25] Limsopatham, Nut, and Nigel Collier. "Learning orthographic features in bi-directional lstm for biomedical named entity recognition." Proceedings of the Fifth Workshop on Building and Evaluating Resources for Biomedical Text Mining (BioTxtM2016). 2016.
[26] Mahmood, ASM Ashique, et al. "DiMeX: a text mining system for mutation-disease association extraction." PloS one 11.4 (2016): e0152725.
[27] Miwa, Makoto, et al. "Wide coverage biomedical event extraction using multiple partially overlapping corpora." BMC bioinformatics 14.1 (2013): 1-12.
[28] Rebholz-Schuhmann, Dietrich, et al. "Automatic extraction of mutations from Medline and cross-validation with OMIM." Nucleic Acids Research 32.1 (2004): 135-142.
[29] Rivandi, Mahdi, John WM Martens, and Antoinette Hollestelle. "Elucidating the underlying functional mechanisms of breast cancer susceptibility through post-GWAS analyses." Frontiers in genetics 9 (2018): 280.
[30] Si, Yuqi, and Kirk Roberts. "A frame-based NLP system for cancer-related information extraction." AMIA Annual Symposium Proceedings. Vol. 2018. American Medical Informatics Association, 2018.
[31] Wei, Chih-Hsuan, Hung-Yu Kao, and Zhiyong Lu. "GNormPlus: an integrative approach for tagging genes, gene families, and protein domains." BioMed research international 2015 (2015).
[32] Vinyals, Oriol, Meire Fortunato, and Navdeep Jaitly. "Pointer networks." arXiv preprint arXiv:1506.03134 (2015).

[33] Beltagy, Iz, Kyle Lo, and Arman Cohan. "SciBERT: A pretrained language model for scientific text." arXiv preprint arXiv:1903.10676 (2019).